\documentstyle[11pt,a4wide]{article}
\begin{document}
\newcommand{\commentout}[1] {}
\newcommand{\DATR}{{\small \sf DATR}}
\newcommand{\PATR}{{\small \sf PATR}}
\newcommand{\LTAG}{{\small \sf LTAG}}
\newcommand{\HPSG}{{\small \sf HPSG}}
\newcommand{\TIC}{{\small \sf TIC}}
\newcommand{\parentn}{{\sf parent}}
\newcommand{\leftn}{{\sf left}}
\newcommand{\rightn}{{\sf right}}
\newcommand{\topn}{{\sf top}}
\newcommand{\bottomn}{{\sf bottom}}
\newcommand{\catn}{{\sf cat}}
\newcommand{\typen}{{\sf type}}
\newcommand{\formn}{{\sf form}}
\newcommand{\IV}{{\small \sf INTRANS\_VERB}}
\newcommand{\TV}{{\small \sf TRANS\_VERB}}
\newcommand{\DIV}{{\small \sf DITRANS\_VERB}}
\newcommand{\DOV}{{\small \sf DOUBLEOBJ\_VERB}}
\newcommand{\AV}{{\small \sf AUX\_VERB}}

\title{\vspace*{-1cm}Using default inheritance to describe
LTAG\footnote{This paper also appeared in
{\em 3e Colloque International sur les
grammaires d'Arbres
Adjoints (TAG+3).}  Technical Report TALANA-RT-94-01, TALANA,
Universit\'e Paris 7, 1994.}}
\author{Roger Evans\thanks{ITRI, University of Brighton, email {\tt
Roger.Evans@itri.bton.ac.uk}, supported by an EPSRC Advanced
Fellowship}, Gerald Gazdar \& David Weir\thanks{COGS, University of
Sussex, email {\tt Gerald.Gazdar/David.Weir@cogs.susx.ac.uk}.}}
\date{September 1994}
\maketitle
\vspace*{-1cm}

\begin{abstract}
We present the results of an investigation into how the set of
elementary trees of a Lexicalized Tree Adjoining Grammar can be
represented in the lexical knowledge representation language \DATR\
(Evans \& Gazdar 1989a,b). The \LTAG\ under consideration is based on
the one described in Abeille {\em et al.} (1990).  Our approach is
similar to that of Vijay-Shanker \& Schabes (1992) in that we formulate
an inheritance hierarchy that efficiently encodes the elementary trees.
However, rather than creating a new representation formalism for this
task, we employ techniques of established utility in other
lexically-oriented frameworks. In particular, we show how \DATR's
default mechanism can be used to eliminate the need for a non-immediate
dominance relation in the descriptions of the surface \LTAG\ entries.
This allows us to embed the tree structures in the feature theory in a
manner reminiscent of \HPSG\ subcategorisation frames, and hence express
lexical rules as relations over feature structures.
\end{abstract}

Vijay-Shanker \& Schabes (1992) have drawn attention to the considerable
redundancy inherent in \LTAG\ lexicons that are expressed in a flat
manner with no sharing of structure or properties across the elementary
trees\footnote{Other related work includes Becker (1993) and Habert
(1991).}. In addition to theoretical considerations, one practical
consequence of such redundancy is that maintenance of the lexicon
becomes very difficult. One way to minimise redundancy is to adopt a
hierarchical lexicon structure with inheritance and lexical rules.
Vijay-Shanker \& Schabes outline such a view of an \LTAG\ lexicon which
is loosely based on that of Flickinger (1987) but tailored for \LTAG\
trees rather than \HPSG\ subcategorization lists.

We share their perception of the problem and agree that adopting a
hierarchical approach provides the best available solution to it.
However, we see no need for the creation of a hierarchical lexical
formalism that is specific to the \LTAG\ problem.  The use of
hierarchical lexicons to reduce or eliminate lexical redundancy is now a
fairly well researched area of NLP (Daelemans \& Gazdar 1992; Briscoe
{\em et al.} 1993) and a variety of formal languages for defining such
lexicons already exist. One of the more widely known and used of these
languages is \DATR\ (Evans \& Gazdar 1989a,b); in this paper we will
show how \DATR\ can be used to formulate a compact, hierarchical
encoding of an \LTAG\ lexicon.

There are three major advantages to using an ``off the shelf''
lexical knowledge representation language (LKRL) like \DATR.  The
first is that it makes it easier to compare the \LTAG\ lexicon with
those associated with other types of lexical grammar. Thus, for
example, \DATR\ has been used to define lexicons for Word Grammar,
\HPSG\ and \PATR-style fragments. The second is that one can take
advantage of existing analyses of other levels of lexical
description. \DATR\ is a general purpose LKRL, not one that is
restricted to syntactic description.  It has been used for phonology,
prosody, morphology, compositional semantics and lexical semantics,
as well as for syntax\footnote{See, for example, Bleiching (1992,
1994), Brown \& Hippisley (1994), Corbett \& Fraser (1993), Cahill
(1990, 1993), Cahill \& Evans (1990), Fraser \& Corbett (in press),
Gazdar (forthcoming), Gibbon (1992), Kilgarriff (1993), Kilgarriff \&
Gazdar (in press), Reinhard \& Gibbon (1991).}. And the third
advantage is that one can exploit existing formal and implementation
work on the language\footnote{See, for example, Andry {\it et al.}
(1992) on compilation, Kilbury {\it et al.} (1991) on coding DAGs,
Langer (1994) on reverse querying, and Barg (1994), Light (1994),
Light {\it et al.} (1993) and Kilbury {\it et al.} (1994) on
automatic acquisition.  And there are at least a dozen different
\DATR\ implementations available on various platforms and programming
languages.}.

Writing an \LTAG\ lexicon in \DATR\ turns out to be little different
from writing any other type of lexicalist grammar's lexicon in an LKRL.
The only significant difference lies in the specification for the
subcategorization frames for lexemes.  In \HPSG\ these are lists of
categories whilst in \LTAG\ they are trees.  In their proposal,
Vijay-Shanker \& Schabes use inheritance to monotonically build up
partial descriptions of trees. Each description is a finite set of
dominance, immediate dominance and linear precedence statements in a
tree description language developed by Rogers \& Vijay-Shanker (1992).
In our approach, trees are described using only local relations: \DATR's
non-monotonicity allows us to ``rewrite'' local relations among the
nodes of a tree, thereby achieving the principal effect of Vijay-Shanker
\& Schabes' use of non-immediate dominance. This allows us to encode tree
relations in the feature structure (in a similar fashion to \HPSG\
subcategorisation lists), and cast lexical rules as relations over
feature structures.

Before going into the details of our account, we note that we have made
a number of simplifying assumptions in our approach to
\LTAG\footnote{And note in addition that the examples {\em in this
paper} are simplified still further for expositional purposes.}. In
particular, our account is purely syntactic, and does not distinguish
\topn\ and \bottomn\ feature specifications on tree nodes, as would be
required to support adjunction and substitution (although it
straightforwardly could do so).

The tree encoding we use is a variant of Kilbury's (1990) bottom-up
encoding of trees.  A tree is described relative to a particular
distinguished leaf node -- in the \LTAG\ entries it is the anchor
node\footnote{Or the principal anchor node, in constructions which have
more than one lexical anchor.}. The tree is represented with the binary
relations \parentn, \leftn\ and \rightn, encoded as features whose
values are the feature structures representing the parent,
immediate-left or immediate-right subtree, encoded in the same way. For
the parent subtree, the distinguished leaf is the parent node
itself\/\footnote{The parent is not a leaf of the entire tree, of
course, but it is a leaf of the subtree {\em above} the current node.},
for non-atomic left and right subtrees, we choose the (generally unique)
lexical leaf of the subtree. Features other than \parentn, \leftn\ or
\rightn\ describe properties of the current node in a conventional
fashion.

As an example, here is the encoding of the subcategorization tree for a
ditransitive verb using a \DATR-style syntax:
{\small
\begin{verbatim}
    Give:
        <cat> == v                          <parent parent cat> == s
        <right cat> == np                   <right right cat> == p
        <parent cat> == vp                  <right right parent cat> == pp
        <parent left cat> == np             <right right right cat> == np.
\end{verbatim}
}

This says that {\sf Give} is a verb, with an NP to its right, a VP as
its parent and an NP to the left of its parent. The second column
stipulates an S as its grandparent and describes the PP to the right of
its right NP, starting at the P, with an NP to its right and PP as its
parent.

Very little of this information is actually specified directly in the entry for
{\sf Give}, however. In fact, {\sf Give} is a minimally specified leaf
of the inheritance hierarchy given by the \DATR\ fragment in
Figure~\ref{fig}\/\footnote{To gain a superficial understanding of this
fragment, read a line such as {\tt <> == INTRANS\_VERB} as {\em
``inherit everything from the definition of \IV.''}, and a line such as
{\tt <parent> == PP\_TREE:<>} as {\em ``inherit the \parentn\ subtree
from the definition of {\small \sf PP\_TREE}.''} Inheritance is always
default -- locally defined feature specifications take priority over
inherited ones.}.
\begin{figure}[htb]
\begin{centering}
\small
\begin{verbatim}
    TREE_NODE:
        <> == undef
        <type> == normal.

    S_TREE:                                 VP_TREE:
        <> == TREE_NODE                         <> == TREE_NODE
        <cat> == s.                             <cat> == vp
    PP_TREE:                                    <parent> == S_TREE:<>
        <> == TREE_NODE                         <left> == NP_TREE:<>.
        <cat> == pp.                        P_TREE:
    NP_TREE:                                    <> == TREE_NODE
        <> == TREE_NODE                         <cat> ==  p
        <cat> == np                             <type> == anchor
        <type> == substitution.                 <parent> == PP_TREE:<>
                                                <right> == NP_TREE:<>.

    INTRANS_VERB:                           DITRANS_VERB:
        <> == TREE_NODE                         <> == TRANS_VERB
        <cat> == v                              <right right> == P_TREE:<>
        <type> == anchor                        <right right root> == to.
        <parent> == VP_TREE:<>.             DOUBLEOBJ_VERB:
    TRANS_VERB:                                 <> == TRANS_VERB
        <> == INTRANS_VERB                      <right right> == NP_TREE:<>.
        <right> == NP_TREE:<>.

    Die:                                    Give:
        <> == INTRANS_VERB                      <> == DITRANS_VERB
        <root> == die.                          <root> == give.
    Eat:
        <> == TRANS_VERB
        <root> == eat.
\end{verbatim}
\caption{DATR Fragment}
\label{fig}
\end{centering}
\end{figure}
Looking at this fragment from the bottom upwards, we see that lexical
entries {\sf Die, Eat, Give} (etc.) are defined in terms of abstract
verb classes \IV, \TV, \DIV, \DOV\ (etc.). These are defined in terms of
each other (so \TV\ is defined as an instance of \IV\ with an additional
NP complement etc.), and ultimately in terms of {\small \sf S\_TREE,
PP\_TREE, NP\_TREE, VP\_TREE, P\_TREE} (etc.) which define fragments of
tree structure. At the very top of the hierarchy {\small \sf TREE\_NODE}
specifies default properties of all nodes in \LTAG\ trees.

This fragment defines by inheritance all the specifications for {\sf
Give} given previously. In addition it defines {\sf root} as a simple
morphological tag on lexical tree-nodes, and \typen, which specifies the
node type as one of {\sf normal, substitution} or {\sf anchor} -- a
classification assumed to have significance for components outside the
lexicon. Nothing precludes more than one leaf in a tree having \typen\
{\sf anchor}. Indeed in the tree for {\sf Give}, the preposition {\sf
to} is also an anchor. But the tree is {\em encoded} from a single leaf,
which we assume to be the `principal' anchor.

A key feature of the analysis is that trees are always fully specified
using only local relations. When a verb class inherits tree information
from a superclass, it can extend, delete or rearrange that information
as required to construct its own tree, overriding tree relations of the
superclass which are not valid in the subclass. This obviates the need
for any kind of partial descriptions of trees required by monotonic
inheritance frameworks. It is this that makes it possible to encode the
trees directly in the feature theory, and this in turn allows us to
encode lexical rules as relations over feature structures.

In the present investigation we have considered four lexical rules:
simple versions of passive, dative, subject-auxiliary inversion (SAI) and
wh-questions (WHQ). Of these, the dative rule is the most
straightforward. We express dative as an explicit alternation for
ditransitive verbs, using the techniques of Kilgarriff (1993). For any
ditransitive, the feature path {\tt<alt dative>} defines a complete
alternative feature structure with a double-object complement. This is
achieved via a single additional statement in the definition of \DIV.
{\small
\begin{verbatim}
    DITRANS_VERB:
        <alt dative> == DOUBLEOBJ_VERB:<>.
\end{verbatim}
}

Passive and SAI are slightly more complicated, because rule application
is `triggered' by the presence of a particular feature in the lexical
entry: {\tt <form> == passive} and {\tt <form> == inv} respectively.
This is achieved in \DATR\ by using the trigger feature to control value
definitions and inheritance. For example, the definition for the class
of auxiliary verbs might be written as follows.
{\small
\begin{verbatim}
    AUX_VERB:                               AUX_TREE:
        <> == INTRANS_VERB                      <> == TREE_NODE
        <aux> == true                           <cat> == <aux_cat "<form>">
        <parent> == AUX_TREE:<>                 <aux_cat> == vp
        <right> == AUX_TREE:<>                  <aux_cat inv> == s.
        <right type> == foot.
\end{verbatim}
}

Here \AV\ defines the (auxiliary) tree for auxiliary verbs, but inherits
the parent and right sister information from {\small \sf AUX\_TREE}. The
latter uses the setting of \formn\ (at the node for the lexical entry
itself -- hence the quotes) to establish the value for \catn: {\sf s}
for inverted verbs, {\sf vp} for all others. Passive is similar, but in
this case the tree structure itself is modified (in the definition of
\TV) when \formn\ is specified as {\sf passive}.

Finally, WHQ has the most complicated treatment. In \LTAG, it is
possible to construe WHQ as a triggered rule like passive and SAI. In
this case, the trigger is the presence of {\tt <form> == null} on {\em
any} NP in the tree. Identifying whether such a trigger is
present in an arbitrary tree requires some fairly subtle \DATR\ code.
However, once the trigger has been located, the effect of this rule is
fairly easy to capture: the top of the verbal subcategorisation tree is
extended by `activating' the following definitions.
{\small
\begin{verbatim}
    INTRANS_VERB:
        <parent parent parent cat> == s
        <parent parent left cat> == np
        <parent parent left form> == wh.
\end{verbatim}
}

The full version of this \DATR\ fragment encodes all the components
discussed above in a single coherent, but slightly more complicated
account. It is available on request from the authors.

\newcommand{\bibentry}[2]%
    {\noindent\hangindent=0.8cm\hangafter=1 {\bf #1} #2\\[-10pt]}
\section*{References}
\small
\bibentry{
Anne Abeille, Kathleen Bishop, Sharon Cote \& Yves Schabes
(1990)}{{\em A lexicalized tree adjoining grammar for English.}
Technical Report MS-CIS-90-24, Department of Computer and Information
Science, University of Pennsylvania.}

\bibentry{
Francois Andry, Norman Fraser, Scott McGlashan, Simon Thornton, \& Nick
Youd (1992)}{Making DATR work for speech: lexicon
compilation in SUNDIAL.  {\it Computational Linguistics} {\bf 18,}
245-267.}

\bibentry{
Petra Barg (1994)}{
Automatic acquisition of DATR theories from observations.
Theories des Lexicons: Arbeiten des Sonderforschungsbereichs 282,
Heinrich-Heine University of Duesseldorf.}

\bibentry{
Tilman Becker (1993)}{
{\it HyTAG: A new type of Tree Adjoining Grammar for hybrid syntactic
representation of free word order languages.} PhD thesis,
Universit\"{a}t des Saarlandes.}

\bibentry{
Doris Bleiching (1992)}{
Prosodisches Wissen in Lexicon.
In G. Goerz, ed.,
{\it Proceedings of KONVENS-92,}
Berlin: Springer-Verlag, 59-68.}

\bibentry{
Doris Bleiching (1994)}{
Integration von Morphophonologie und Prosodie in ein hierarchisches Lexicon.
To appear in {\it Proceedings of KONVENS-94.}}

\bibentry{
Ted Briscoe, Valeria de Paiva \& Ann Copestake, eds. (1993)}{
{\it Inheritance, Defaults, and the Lexicon,}
Cambridge: Cambridge University Press.}

\bibentry{
Dunstan Brown \& Andrew Hippisley (1994)}{
Conflict in Russian genitive plural assignment: A solution
represented in DATR.
{\it Journal of Slavic Linguistics,} {\bf 2(1),} 48-76.}

\bibentry{
Greville Corbett \& Norman Fraser (1993)}{  Network Morphology:
a \DATR\ account of Russian nominal inflection. {\it Journal
of Linguistics} {\bf 29,} 113-142.}

\bibentry{
Lynne Cahill (1990)}{ Syllable-based morphology.  {\it COLING-90},
Vol. 3, 48-53.}

\bibentry{
Lynne Cahill (1993)}{  Morphonology in the lexicon.
{\it Sixth Conference of the European Chapter of the Association
for Computational Linguistics,} 87-96.}

\bibentry{
Lynne Cahill \& Roger Evans (1990)}{An application of \DATR: the
\TIC\ lexicon,
in  {\it Proceedings of the 9th European Conference on Artificial
Intelligence,} Stockholm, 120-125.}

\bibentry{
Walter Daelemans \& Gerald Gazdar, eds. (1992)}{
{\it Computational Linguistics} {\bf 18.2} \& {\bf 18.3}, special issues
on inheritance.}

\bibentry{
Roger Evans \& Gerald Gazdar (1989a)}{ Inference in \DATR .
{\it Fourth Conference of the European Chapter of the
Association for Computational Linguistics,} 66-71.}

\bibentry{
Roger Evans \& Gerald Gazdar (1989b)}{ The semantics of \DATR .
In Anthony G. Cohn, ed.  {\it Proceedings of the Seventh
Conference of the Society for the Study of Artificial
Intelligence and Simulation of Behaviour.}  London:
Pitman/Morgan Kaufmann, 79-87.}

\bibentry{
Daniel P. Flickinger (1987)}{ {\it Lexical Rules in the
Hierarchical Lexicon}, doctoral dissertation, Stanford
University.}

\bibentry{
Norman Fraser \& Greville Corbett (in press)}{
Gender, animacy, and declensional class assignment: a unified account
for Russian.  {\it Year Book of Morphology 1994.}}

\bibentry{
Gerald Gazdar (forthcoming)}{
Ceteris paribus.
To appear in J.A.W. Kamp \& C. Rohrer, eds.
{\it Aspects of Computational Linguistics.}}

\bibentry{
Dafydd Gibbon (1992)}{  ILEX: a linguistic approach to computational
lexica.  In Ursula Klenk, ed. {\it Computatio Linguae: Aufsaze zur
algorithmischen und quantitativen Analyse der Sprache (Zeitschrift
fur Dialektologie und Linguistik,} Beiheft 73),
Stuttgart: Franz Steiner Verlag, 32-53.}

\bibentry{
Beno\^{\i}t Habert (1991)}{
Using inheritance in object-oriented programming to combine syntactic
rules and lexical idiosyncrasies.
{\it Proceedings of the second International Workshop on Parsing
Technologies,} Cancun, Mexico.}

\bibentry{
James Kilbury (1990)}{
{\it Encoding constituent structure in feature structures,}
unpublished manuscript, University of
Duesseldorf.}

\bibentry{
James Kilbury, Petra Naerger \& Ingrid Renz (1991)}{
\DATR\ as a lexical component for \PATR. {\it Fifth Conference of the
European Chapter of the Association for Computational Linguistics,}
137-142.}

\bibentry{
James Kilbury, Petra Barg \& Ingrid Renz (1994)}{
Simulation lexicalischen Erwerbs
In Sascha W. Felix, Christopher Habel \& Gert Rickheit
{\it Kognitive Linguistik: Repraesentation und Prozesse.}
Opladen: Westdeutscher Verlag, 251-271.}

\bibentry{
Adam Kilgarriff (1993)}{  Inheriting verb alternations.
{\it Sixth Conference of the European Chapter of the Association
for Computational Linguistics,} 213-221.}

\bibentry{
Adam Kilgarriff \& Gerald Gazdar (in press)}{
Polysemous relations.
In F.R. Palmer, ed. {\it Festschrift for John Lyons.}
Cambridge: Cambridge University Press, 1994.}

\bibentry{
Hagen Langer (1994)}{ Reverse queries in \DATR.
{\it COLING-94}}.

\bibentry{
Marc Light (1994)}{
Classification in feature-based default inheritance hierarchies
To appear in {\it Proceedings of KONVENS-94.}}

\bibentry{
Marc Light, Sabine Reinhard \& Marie Boyle-Hinrichs (1993)}{
INSYST: an automatic inserter system for hierarchical lexica.
{\it Sixth Conference of the European Chapter of the Association for
Computational Linguistics,} 471.}

\bibentry{
Sabine Reinhard \& Dafydd Gibbon (1991)}{
Prosodic inheritance and morphological generalisations.
{\it Fifth Conference of the European Chapter of the Association
for Computational Linguistics,} 131-136.}

\bibentry{
J. Rogers \& K. Vijay-Shanker (1992)}{
Reasoning with descriptions of trees.
{\it 30th Annual Meeting of the Association for
Computational Linguistics,} 72-80.}

\bibentry{
K. Vijay-Shanker \& Yves Schabes (1992)}{
Structure sharing in lexicalized tree-adjoining grammars.
{\it COLING-92} Vol. I, 205-211.}
\end{document}